# AI Driven Road Maintenance Inspection

**Ratnajit Mukherjee* (ratnajit.mukherjee@navinfo.eu), Haris Iqbal, Shabbir Marzban, Ahmed Badar, Terence Brouns, Shruthi Gowda, Elahe Arani and Bahram Zonooz**

Advanced Research Lab, Navinfo Europe, Eindhoven, Netherlands

**Abstract**

Road infrastructure maintenance inspection is typically a labour-intensive and critical task to ensure the safety of all the road users. In this work, we propose a detailed methodology to use state-of-the-art techniques in artificial intelligence and computer vision to automate a sizeable portion of the maintenance inspection subtasks and reduce the labour costs. The proposed methodology uses state-of-the-art computer vision techniques such as object detection and semantic segmentation to automate inspections on primary road structures such as the road surface, markings, barriers (guardrails) and traffic signs. The models are mostly trained on commercially viable datasets and augmented with proprietary data. We demonstrate that our AI models can not only automate and scale maintenance inspections on primary road structures but also result in higher recall compared to traditional manual inspections.

## 1. Introduction

Road infrastructure maintenance is traditionally a labour-intensive industry whereby a major portion of the cost involves the proverbial "boots on the ground" to manually check and periodically maintain a stretch of road or a zone. This constant manual intervention leads to high costs and delays in road upkeep. The high costs can however be brought down by a significant margin by introducing the latest computer vision (CV) and artificial intelligence (AI) based technologies to largely automate the process of extracting valuable road damage information from image data and find the most pressing problems.

In this work, we present an approach employing state-of-the-art CV and AI algorithms such as object detection and semantic segmentation to successfully automate the process of accurately detecting the type and the extent of road and surface damages along with their precise geographic locations in GPS coordinates. We have used commercially viable datasets to train and optimize our AI models and collected a small proprietary dataset with varied scenarios for the purpose of testing our solutions in the real world. Details are presented in Section 3 and 4.

AI driven Road Damage Detection

We conclude that the deployment of such modern AI solutions will not only be a key component to significantly lower the labour costs but also increase the efficiency of maintenance based on findings of this system.

## 2. Background

For this case study, we have used two extensively researched AI techniques involving Convolution Neural Networks (CNNs) i.e., *object detection* and *semantic segmentation.* In this section, we provide a brief introduction to these techniques.

*Object Detection*

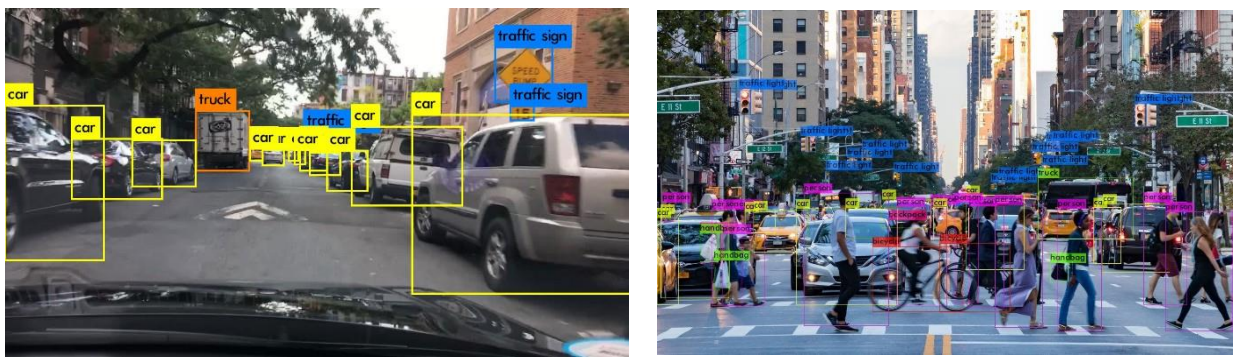

Figure 1: Example output of a generic object detection algorithm.

Object Detection refers to identifying the existence of an object instance and its location in an image. Specifically, for a given image or video frame, along with predefined set of relevant object types, this task aims to predict the location of all the object instances within the image using a bounding box and classify the category of the objects present in them.

A detailed literature on object detection is out of scope of this paper but excellent reviews and detailed discussion on this topic can be found in [1-2]. We used a well-known and robust object detection head, the Single Shot Multibox Detector (SSD) [3], with a state-of-the-art memory-efficient network Harmonic DenseNet (HarDNet) as feature extractor [4] for all our detection tasks. Figure 2 provides a schematic diagram of the feature extractor and detection head.

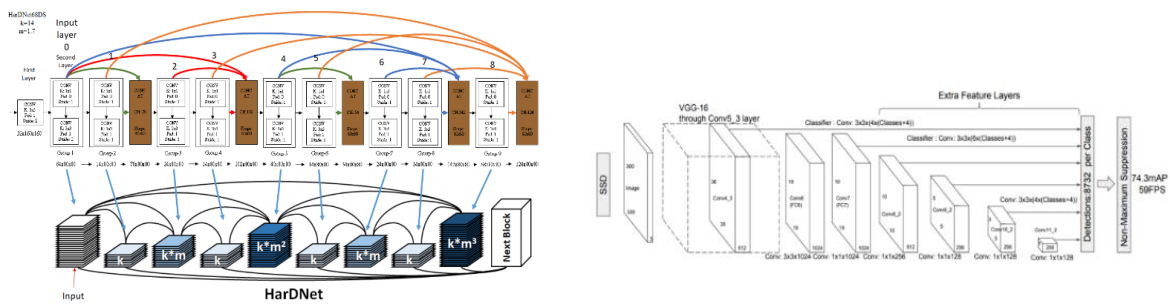

Figure 2: HarDNet backbone [4] (left) with SSD detection head [3] (right).

AI driven Road Damage Detection

*Semantic Segmentation*

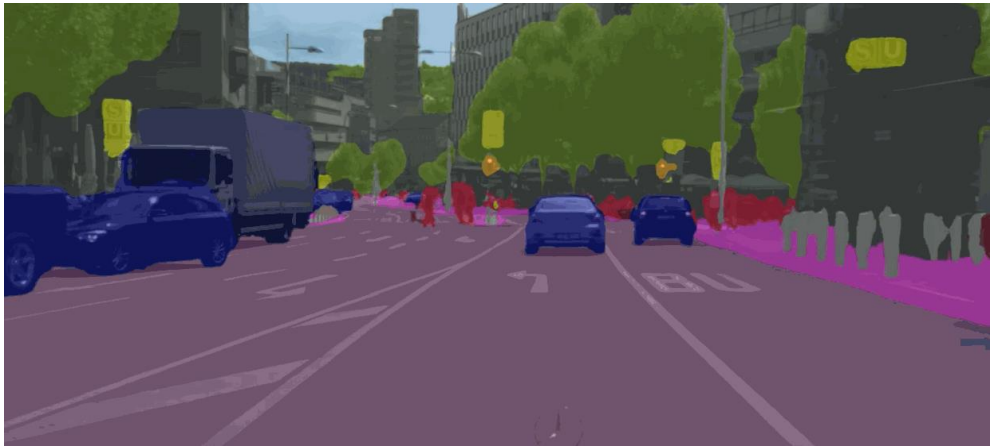

**Figure 3: Example of Semantic Segmentation in a crowded scene.**

In Semantic Segmentation, also known as pixelwise classification, each pixel of an image is assigned a predefined object category, such as roads, lanes and curbs, based on semantic correspondence. A detailed literature on semantic segmentation is out of scope of this paper, but reviews on this topic are available in [5-6].

For semantic segmentation, we have an in-house state-of-the-art deep neural network, RGPNet [7], with a lightweight asymmetric encoder-decoder structure for fast and efficient inference. It comprises of three components: an encoder which extracts high level semantic features, a light asymmetric decoder and an adaptor which links different stages of encoder and decoder. Figure 4 provides a schematic diagram of the segmentation head architecture. Each row has the same spatial resolution with number of channels mentioned in rectangular boxes. The backbone block represents the encoder. The adaptor reduces the number of channels of the encoder output by a factor of 4 using pointwise convolution followed RGPNet segmentation head. The details of this network are available in [7].

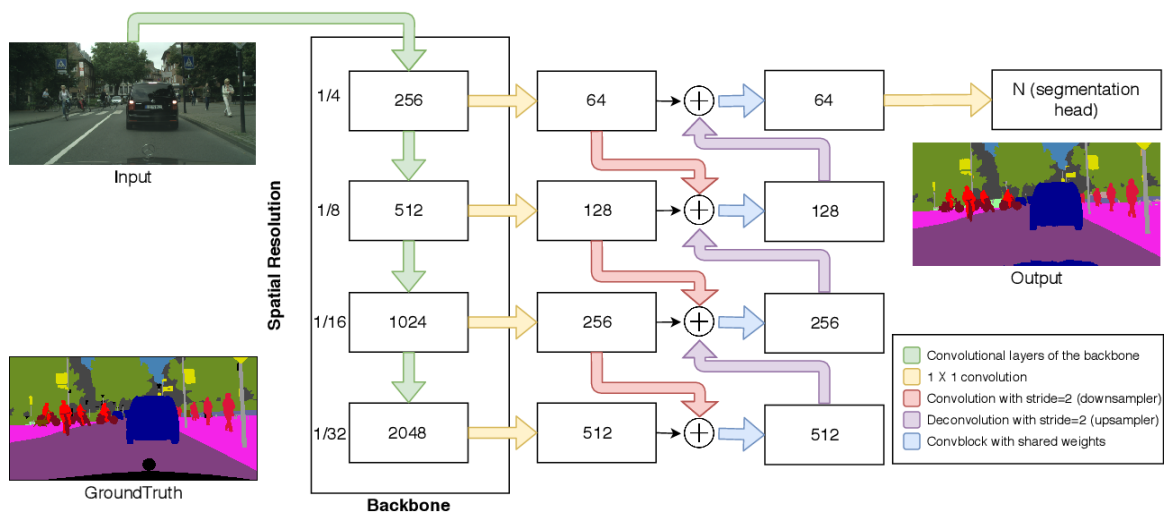

**Figure 4: RGPNet architecture diagram [7].**

AI driven Road Damage Detection

## 3. Approach

In this work, using outputs of the previously introduced tasks, we primarily focus on detecting damages in road surface, road markings and traffic signs with indication of type and extent of damage. We also present solutions to indicate safety of barriers alongside the roads. We selected these subtasks based on their criticality in terms of, safety for road users and their extensive manual labour requirements. This can easily be extended to a larger set of maintenance tasks.

*Road damage detection (road cracks and erosions)*

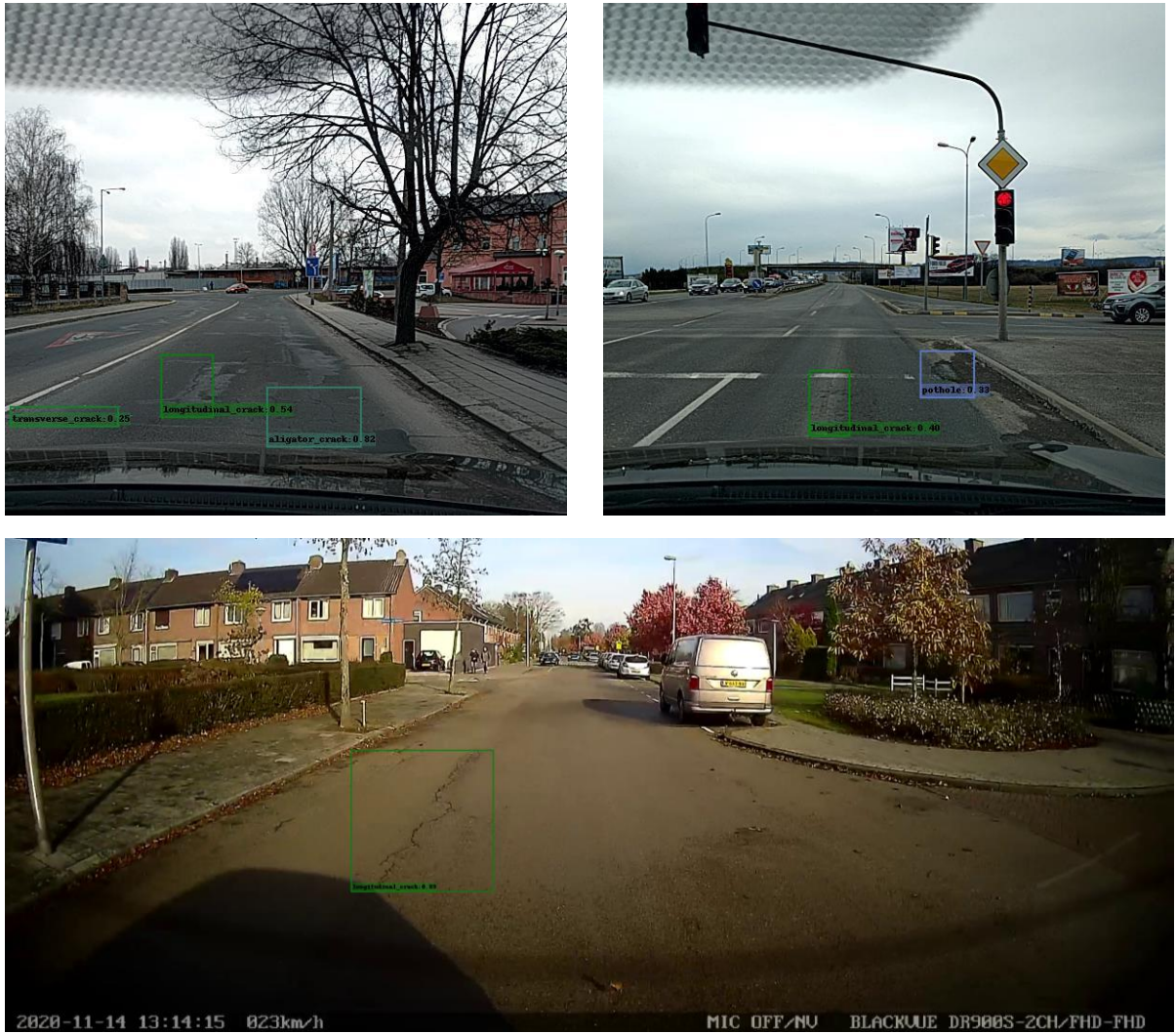

**Figure 5: Examples of road damage detection using our detection model. The damages are marked with colored bounding boxes, with each color indicating a different type of damage.**

Road damage detection has been an active research problem in AI for quite some time and international competitions have also been held for accurate detection of road damages and prior research work on this topic can be found in [8]. Although this problem can be solved using either object detection or semantic segmentation, prior research [9-10] has tackled this challenge using object detection.

AI driven Road Damage Detection

Dataset preparation: Following prior research works, we explored the well-known publicly available Road Damage Dataset (RDD) [8] which consists of road damages from India, Czech Republic and Japan. The variety of road structures and corresponding damages available in this dataset makes it ideal for training robust state-of-the-art models which are expected to perform well even for out-of-distribution data. The dataset consists of multiple categories which we categorized into a few super-categories: *alligator cracks, transverse cracks, longitudinal cracks, missing markings* and *potholes*. We chose the 2019 edition of the dataset to train our model, where 10 percent of the images were randomly selected for the validation set.

Model training: We trained on RDD using a high accuracy version of the HarDNet feature extractor (HarDNet85) with the SSD detection head. The model achieved an accuracy of 60.11 mAP on the validation set with a throughput of 85 frames/sec (FPS) on an Nvidia GTX 2080Ti.

In Figure 5, we visualize the results of the trained road damage detection model on a few images of road scenes. The detected road damages are indicated by colored bounding boxes, where transverse and longitudinal tracks are shown in green, alligator cracks in cyan and potholes in blue.

*Road marking damage*

For road markings, the goal is to detect road marking damage with prediction of the extent of the detected damage in terms of affected area. As one of the primary challenges of this task is the large variety of road markings seen across multiple countries, we focussed on a few pre-defined and common road markings seen across multiple countries such as solid lines, dashed lines, stop lines, arrows, split lines, and road merges. This part of the solution was addressed using our in-house semantic segmentation network, RGPNet.

Dataset preparation: we chose the commercial version Mapillary Vistas [11] dataset whose primary characteristics is its high diversity with street images from multiple countries. The dataset consists of 25000 images from around the world with 152 segmentation categories and the diverse nature of the dataset indirectly ensures the robustness of the trained models in real-world scenarios.

AI driven Road Damage Detection

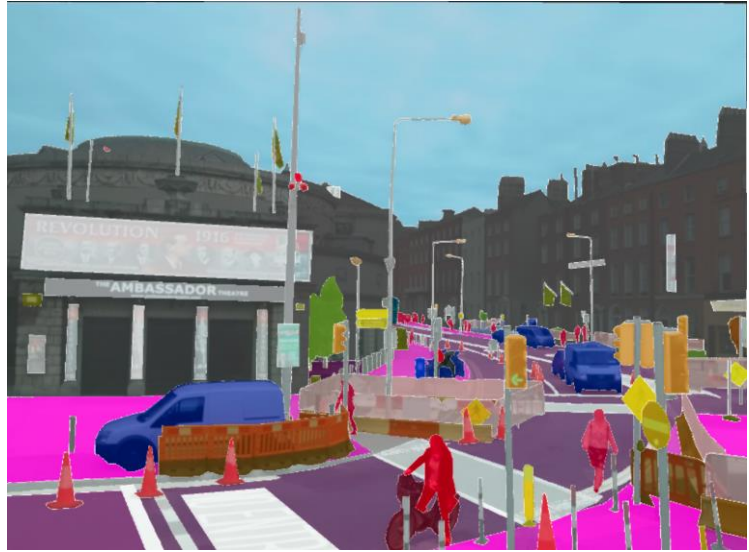

Figure 6: Example from Mapillary Vistas [11] dataset overlaid with the segmentation map.

Model training: We trained a large and accurate model with WiderResnet38 [12] feature extractor and the RGPNet segmentation head on the commercial version of Mapillary Vistas dataset. The model was able to achieve an mIoU of 0.52 on the validation set.

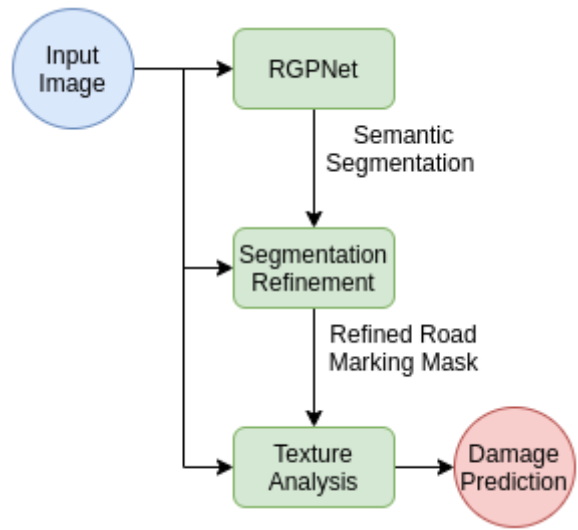

Figure 7: Schematic diagram of the post-processing steps for road marking damage.

Post Processing: For this task, as illustrated in Figure 7, we start with refining the segmentation masks (related to road markings) obtained from RGPNet by using an adaptive pixel intensity thresholding. Refined masks are then fed to a bank of differential filters that give high response on noisy texture which is indicative of marking damage. To have a measure of extent of damage, SLIC [13] super pixels are computed to get density of the response. This output is finally thresholded to indicate damage regions. In Figure 8, we demonstrate an example input and output with key intermediate steps from where damaged markings were extracted using the procedure described earlier. Top left, we show the input image. The

AI driven Road Damage Detection

middle image is the raw output from RGPNet. Top right, we show refined road marking classes. Bottom left image shows the SLIC contours on the region of interest. Bottom middle image shows the output of texture analysis based on the edges present within the region of interest, which is then thresholded and shown in bottom right.

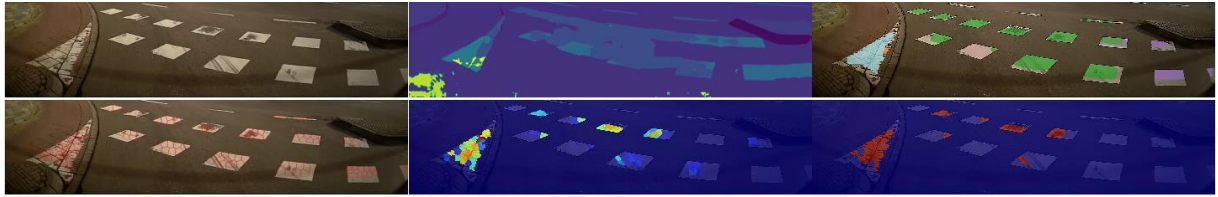

Figure 8: Example of damaged marking analysis.

*Traffic Sign Damage:*

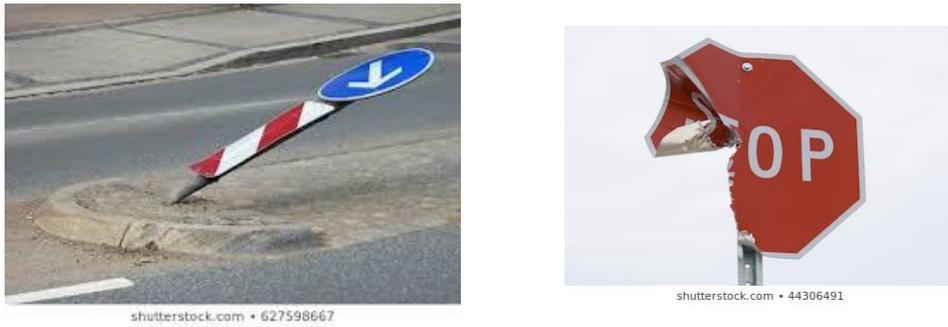

Figure 9: Examples of damaged road signs.

Traffic signs are an important attribute of the road and contribute to road safety but can suffer from damages such as the ones shown in Figure 10, which decrease their visibility to road users. These damages should be quickly detected and repaired to prevent road accidents. We propose to detect damages to traffic signs by training a highly accurate object detection model (as introduced in Section 2) on annotated traffic sign data from a proprietary dataset. Although this training set was collected in a different country, inference results on data from the Netherlands are still accurate (see Figure 11) since the characteristics of traffic signs are similar across the world. For each detected traffic sign, we extract the corresponding bounding box from the image and compare it with a database of non-damaged traffic sign crops of the same type to determine whether the sign is damaged or not.

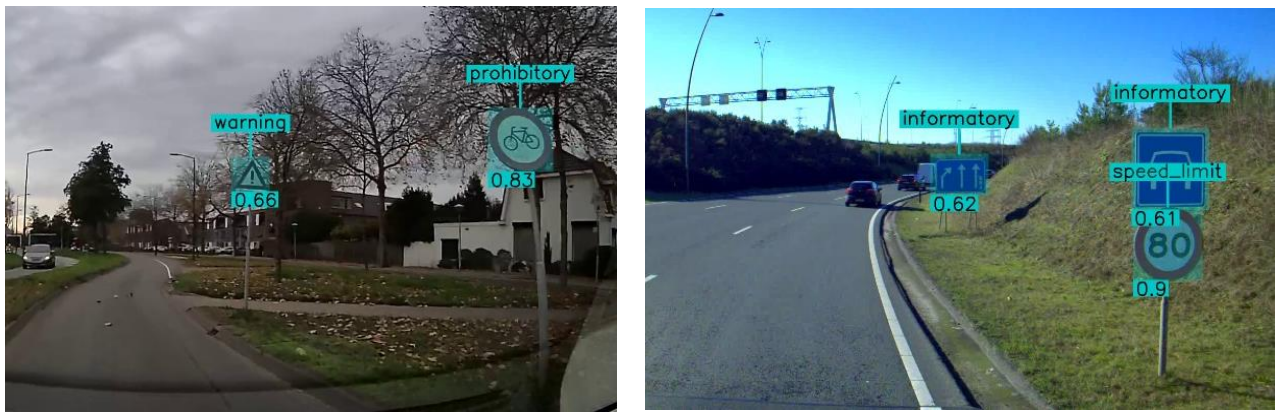

Figure 10: Inference examples from the trained traffic sign detector on out-of-distribution data from the Netherlands.

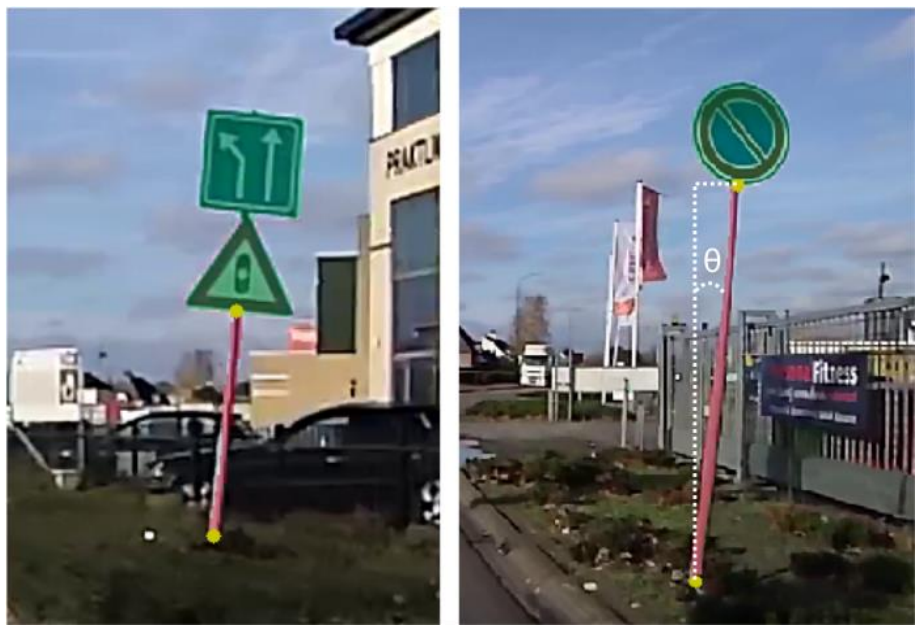

Figure 11: Example of skew calculation for two traffic signs. The predicted segmentation masks for the traffic sign board and corresponding pole are shown in green and red, respectively. In the right image, the skew is given as the angle θ and indicated using the white dotted lines.

To determine whether a traffic sign is skewed, we use RGPNet to segment the traffic sign pole and compare the angle of the segmented pole area with that of the vertical axis of the image plane. Figure 12 shows an example of a skewed traffic sign and the process of determining the damage due to the skew. The angle is determined by finding the minimum rotated rectangle [14] of the pole and then calculating the distance between the middle point of the edge closest and furthest away from the traffic sign centroid (shown as yellow points).

*Barrier Segmentation and Safety Check:*

AI driven Road Damage Detection

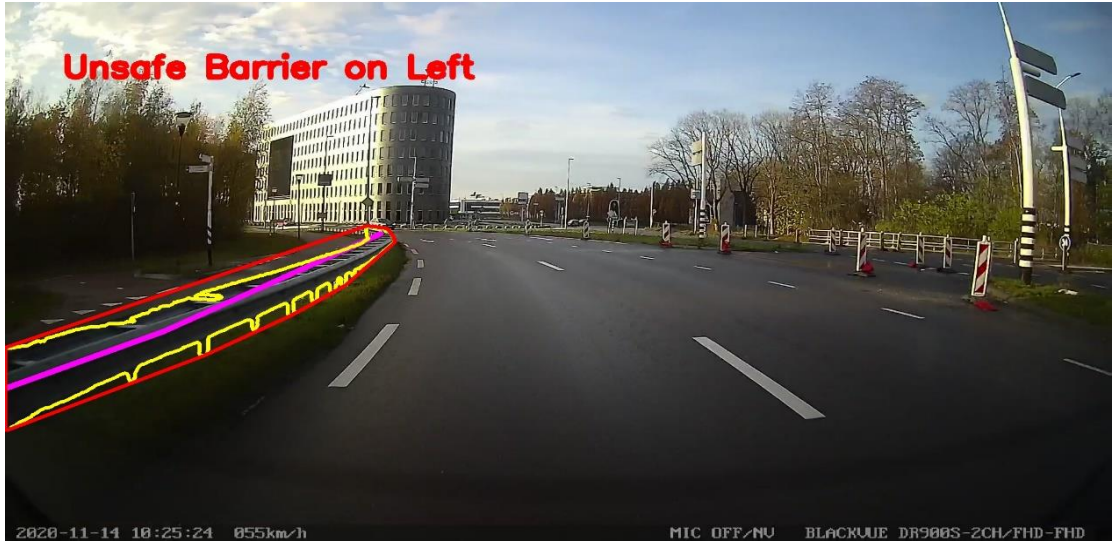

**Figure 12: Example of barrier segmentation and barrier safety check.**

The goal of this subtask was to extract and determine whether the guardrails on the side of the road are motorcycle friendly. The criterion for barrier safety is as follows: a barrier is not motorcycle friendly if there is no shield beneath the guard rail as it can cause serious injury in case of a skid. To that end, we detect the left and right barriers in the scene using RGPNet trained on Mapillary Vistas Dataset. Subsequently, we determine the safety of the guardrails by using the Satoshi algorithm [15] to obtain the contours of the barriers from segmentation masks and then use Sklansky's algorithm [16] to get the convex hull of each contour. We determine the solidity of each contour using the Convex Hull Solidity formula given in Equation (1).

$$\text{Solidity} = \frac{\text{ContourArea}}{\text{ConvexHullArea}} \quad (1)$$

If the solidity of a given contour is larger than a specific threshold then its respective barrier is labelled as safe, otherwise it is labelled as an unsafe barrier. Since the capture viewpoint is from the right (driving side in Europe), we pick the solidity threshold of 0.8 for barriers on the right and 0.6 for those on the left.

## 4. Map Visualization

For practical usage, we aggregated the solutions to each of the subtask to a visualization solution. First, we extract the GPS information from the captured dashcam videos and then, interpolate it to gain GPS coordinates for each frame of the test video. The damaged road markings, erosions, barriers and traffic signs are marked for the specific frames and matched with their corresponding GPS coordinates. In the end, we pin the locations and types of the predicted damages on the map as shown in Figure 13.

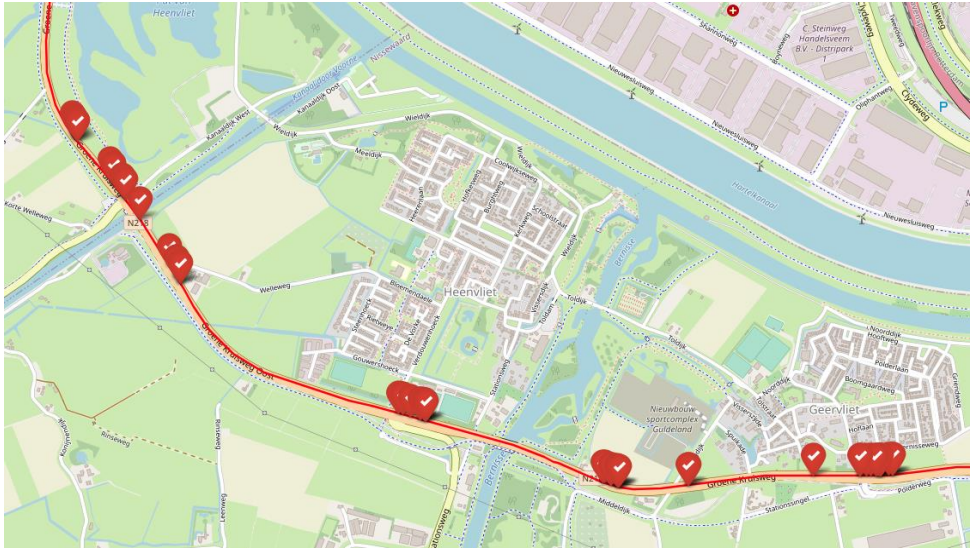

Figure 13: Precise geographical locations of the damages.

## 5. Conclusion

In this work, we show that a sizeable portion of road maintenance inspection can be automated using AI. Furthermore, it can be easily extended to other inspection subtasks with availability of the labelled data. Such automation allows seamless scaling of the solution and results in higher recall rate compared to traditional manual approaches which leads to improved and targeted maintenance.

To improve on localization of the detected inspection items, other modalities such as Lidar in combination with SLAM techniques can be explored. With SLAM, a detailed point cloud representation of entire scene can be reconstructed which can further enhance and extend detection of inspection subtasks.